\documentclass[conference]{IEEEtran}
\usepackage{cite}
\usepackage{amsmath,amssymb,amsfonts}
\usepackage{algorithmic}
\usepackage{graphicx}
\usepackage{textcomp}
\usepackage{xcolor}
\usepackage{multirow}
\usepackage{multicol}
\usepackage{caption}
\usepackage{subcaption}
\usepackage{graphicx}
\usepackage{booktabs}
\usepackage{dblfloatfix}
\newcommand{\RR}{\mathbb{R}}
\newcommand{\cC}{\mathcal{C}}
\newcommand{\cK}{\mathcal{K}}

\newcommand{\cT}{\mathcal{T}}
\newcommand{\cE}{\mathcal{E}}
\newcommand{\SPPA}{S_{\mathrm{PPA}}}

\begin{document}

\title{A PPA-Driven 3D-IC Partitioning Selection Framework with Surrogate Models
}

\author{
    \IEEEauthorblockN{
        Shang Wang$^1$, 
        Shuai Liu$^1$, 
        Owen Randall$^1$,
        Matthew E.~Taylor$^{1,2}$
    }
    \vspace{0.15cm} 
    \IEEEauthorblockA{
        $^1$University of Alberta, Canada\\
        $^2$Alberta Machine Intelligence Institute (Amii), Canada\\
        \texttt{\{shang8,shuai14,davidowe,mtaylor3\}@ualberta.ca}
    }
}

\maketitle

\begin{abstract}
3D-IC netlist partitioning is commonly optimized using proxy objectives,
while final PPA is treated as a costly evaluation rather than an optimization signal.
This proxy-driven paradigm makes it difficult to reliably translate additional PPA evaluations into better PPA outcomes.
To bridge this gap, we present DOPP (\underline{D}-\underline{O}ptimal \underline{P}PA-driven \underline{p}artitioning selection), an approach that bridges the gap between proxies and true PPA metrics.
Across eight 3D-IC designs, our framework improves PPA over Open3DBench (average relative improvements of 9.99\% congestion, 7.87\% routed wirelength, 7.75\% WNS, 21.85\% TNS, and 1.18\% power).
Compared with exhaustive evaluation over the full candidate set, DOPP achieves comparable best-found PPA while evaluating only a small fraction of candidates, substantially reducing evaluation cost. 
By parallelizing evaluations, our method delivers these gains while maintaining wall-clock runtime comparable to traditional baselines.

\end{abstract}

\begin{IEEEkeywords}
3D IC Partitioning, Linear regression, D-optimal design, Power Performance Area (PPA)
\end{IEEEkeywords}
\begin{figure*}[!t]
    \centering
    \includegraphics[width=1.0\linewidth]{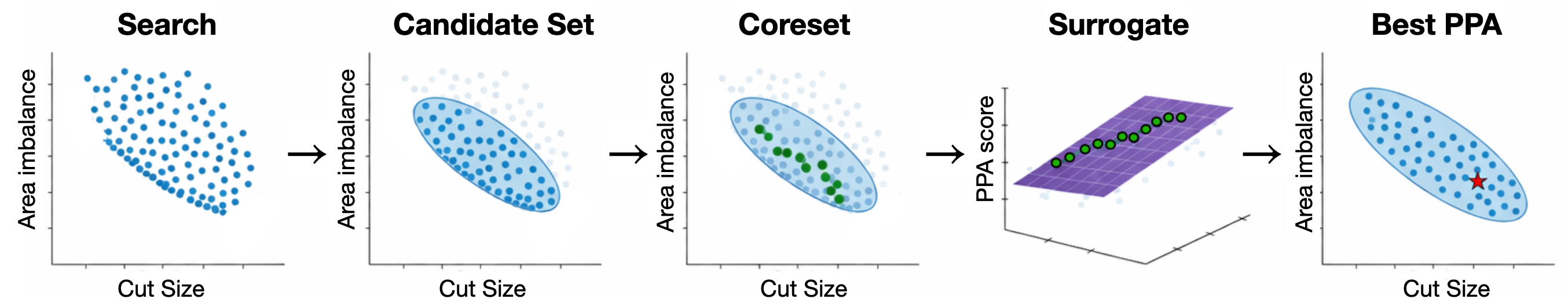}
    \caption{In the DOPP pipeline, search generates a proxy-diverse candidate set, D-optimal selects an informative coreset, a surrogate model is fit to the labeled coreset to rank candidates, and the best PPA solution is selected from the evaluated set.}
    \label{fig:pipeline}
\end{figure*}
\section{Introduction}
\label{introduction}
3D integrated circuits (3D-ICs) have emerged as a promising way to continue performance scaling by vertically stacking multiple dies and leveraging dense vertical interconnections.
While 3D integration can shorten critical interconnections and improve integration density, it also introduces new design challenges~\cite{3DIC}. 
In 3D-IC design flow~\cite{cascade2d, compact2d, shrunk2d,pin3d},
netlist partitioning is a critical early-stage decision that assigns modules or cells to different tiers from a gate-level netlist.

Existing 3D-IC partitioning approaches generally fall into two categories. 
First, traditional methods typically rely on graph/hypergraph partitioning~\cite{KL, FM, MLFM, ababei2005placement, schlag2023high,chipletpart}, simulated annealing~\cite{simulated_annealing,sawicki2009cells}, and  heuristics~\cite{memoryonlogic, ILP, placement_driven_partitioning, BEOL} to iteratively optimize proxy objectives such as cut size, area balance, etc.
Second, machine learning methods~\cite{tpgnn,ta3d} often improve partitioning quality by incorporating richer task-related information into the model.
Both categories iteratively update solutions at the front-end of the flow to optimize proxy metrics, while the final physical metrics of interest, PPA (performance, power, and area), are typically obtained during evaluation, rather than being used as a learning or optimization signal.

However, the success of supervised learning often depends on having a large amount of labeled data, which is difficult to obtain given that PPA evaluation is expensive.
Recent work~\cite{deeplayout, ML_prediction, routing_prediction} has explored training surrogate models (i.e., computationally efficient approximations of complex physical simulations) to generalize over different chips, replacing expensive PPA evaluations, but their effectiveness is limited by data scarcity, the inherent complexity of chip data, and large differences between designs, leading to unreliable generalization.

Motivated by these challenges, we propose a budget-efficient framework that enables direct PPA-driven selection for 3D-IC partitioning.
Our goal is to maximize the final PPA metrics by selecting the most informative candidates for evaluation, while reducing the number of times evaluating PPA where we must invoke the full 3D backend implementation.
A key observation is that when the optimization loop is not explicitly coupled with PPA, increasing compute budget tends to yield better performance in the proxy space, but does not reliably translate into PPA gains. 
Therefore, the central question we address is: \textit{\textbf{how can we allocate a small number of expensive PPA evaluations so that additional compute budget can be converted into consistent improvements in PPA metrics?}}

To address this, we propose DOPP (D-Optimal PPA-driven Partitioning selection), illustrated in Fig.~\ref{fig:pipeline}, to explicitly align early-stage proxy optimization with true PPA metrics. 
DOPP first employs an annealing-based search to generate a diverse set of candidates near the Pareto frontier. 
Then a coreset (i.e., a subset of all the candidates) is constructed to cover the most informative candidates.
Next, DOPP evaluates PPA values by running full physical simulations on the coreset, which are later used to fit a local surrogate model on PPA scores. 
The PPA scores predicted by the surrogate model are used to rank the full candidate set.
Lastly, DOPP evaluates a small set of top-ranked candidates, obtaining PPA values.
The final solution is chosen among the union of the coreset and the surrogate's top-ranked candidates.

DOPP effectively learns a surrogate from the candidate set with modest computational resources, identifying solutions with high-performing PPA metrics. Our empirical results show DOPP effectively bridges the proxy-to-PPA gap; furthermore, it contributes a flexible selection methodology that enables users to retrieve solutions tailored to their specific PPA preferences.

\section{Framework}
\label{framework}
We adopt the 3D-IC design workflow and evaluation framework provided by Open3DBench~\cite{open3dbench} throughout this work.
Open3DBench provides the first fully open-source-tool-based 3D backend implementation and PPA evaluation benchmark targeting face-to-face (F2F) hybrid-bonded 3D-ICs. 

In Open3DBench, partitioning is studied under a memory-on-logic setting: all logic cells are assigned to the bottom tier, while memory blocks are distributed between the top and bottom tiers. 
Partitions are explored using simulated annealing (SA) under a fixed search budget. 
The SA objective is a fixed weighted combination of (i) cross-tier cut size and, representing the total number of vertical interconnections; and (ii) inter-tier area utilization imbalance, which measures the area disparity between tiers.

Our framework, DOPP, operates in two stages: Diverse Candidate Set Generation (Section~\ref{diverse candidate set generation}) and Local Surrogate Modeling with D-Optimal Design (Section~\ref{linear surrogate model and d-optimal design}). 
Unlike Open3DBench and prior work that target a single proxy-optimal solution, DOPP generates a diverse candidate pool and retrieves the best PPA solution tailored to user preferences with modest evaluation cost.

\subsection{Diverse Candidate Set Generation}
\label{diverse candidate set generation}
\begin{figure*}[t]
    \centering
    \includegraphics[width=1.0\linewidth]{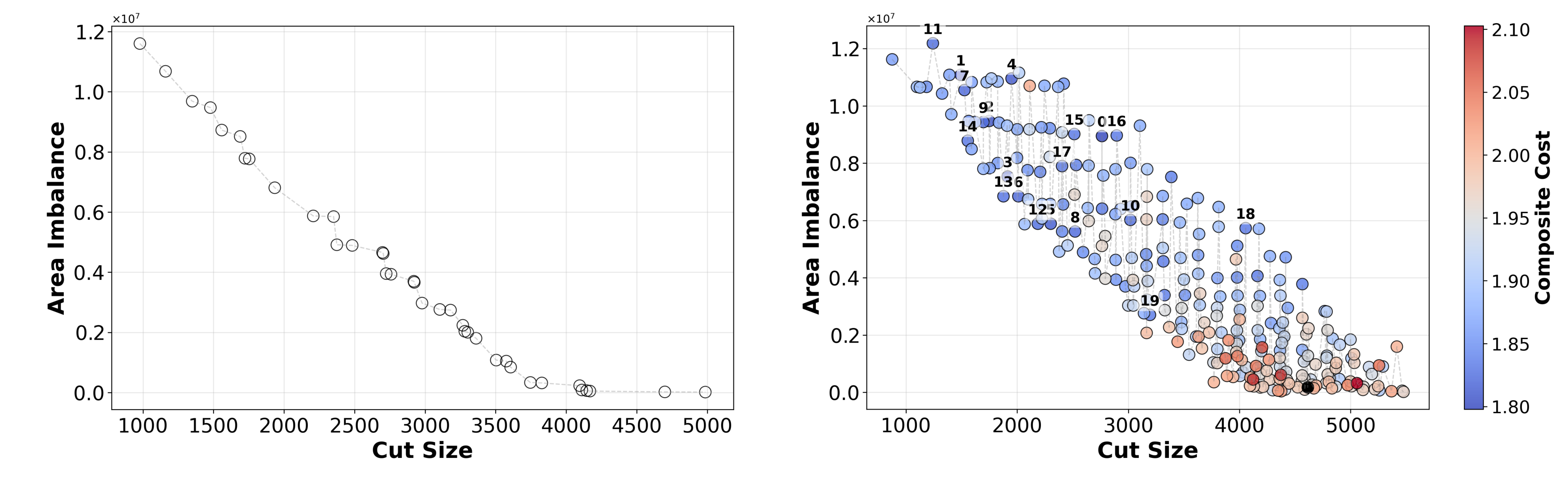}
    \caption{Proxy-space candidate sets under two archive strategies on \texttt{bp\_multi}. Left: vanilla multi-objective archive that keeps only non-dominated solutions, producing a small proxy frontier (47 candidates). Right: grid-based proxy archive that retains at most one proxy-competitive representative per grid cell, yielding a much denser and more diverse set (257 candidates), including solutions behind the frontier. Points are colored by PPA composite cost (lower is better), and numeric labels denote the candidate’s rank under this cost.}
    \label{fig:pareto}
\end{figure*}

To ultimately select a high quality design, we first require a candidate set covering solutions with strong PPA performance. 
We assume that proxy metrics are a useful but imperfect indicator of PPA: proxies and PPA are roughly positively correlated, yet a proxy-optimal solution is not necessarily PPA optimal.
Therefore, we generate a diverse set of candidates with good proxy metric values to increase the chance of finding candidates with strong PPA performance.
We build upon the simulated annealing procedure and extend it to a multi-objective setting by maintaining a grid-based Pareto archive over the proxy space. 
Over the full annealing trajectory, we explore many candidate moves, and the set of solutions retained in the grid-based archive constitutes the final candidate set.
We adopt a grid-based archive for two reasons: (i) the proxy Pareto frontier alone can be small and concentrated, limiting diversity across trade-off regions, and (ii) candidates that are dominated in proxy space can still achieve strong PPA and should not be discarded, seeing Fig.~\ref{fig:pareto}.
Concretely, we log-normalize the proxy objectives and discretize them into a fixed 2D grid, keeping at most one representative candidate per grid cell. 
Within each cell, we maintain a non-dominated representative by replacing the stored solution if a new candidate is at least as good in both proxy metrics. 
This archive retains not only Pareto optimal solutions but also a subset of proxy metric dominated, yet still competitive, solutions.
Section~\ref{sec:candidate} empirically validates the misalignment assumption between proxy and PPA and analyzes how the grid resolution affects candidate diversity and downstream PPA performance.


PPA quality is summarized by an aggregated PPA composite cost.
Let $\mathcal{C}$ denote the candidate set and let $m_k(c)$ be the $k^{th}$ PPA metric of candidate $c\in\mathcal{C}$ for $k \in \{1,...,K\}$, where $K$ is the total number of PPA metrics.
We map each $m_k(c)$ in $\mathcal{C}$ to a normalized metric $\tilde{m}_k(c)$ using min-max scaling as follows:
\begin{equation}
\tilde{m}_k(c)=\frac{m_k(c)-\min_{c'\in\mathcal{C}} m_k(c')}{\max_{c'\in\mathcal{C}} m_k(c')-\min_{c'\in\mathcal{C}} m_k(c')}.
\end{equation}
The PPA composite cost $S_{PPA}$ is then defined as the $\ell_2$-norm of the normalized metric vector $\tilde {\mathbf{m}}(c)=[\tilde m_k(c)]_{k=1}^K\in \RR^K$:
\begin{equation}
S_{\mathrm{PPA}}(c)=\left\lVert \tilde{\mathbf m}(c)\right\rVert_2
= \sqrt{\sum_{k=1}^{K}\tilde{m}_k(c)^2}.
\end{equation}
Smaller $S_{\mathrm{PPA}}$ indicates better overall PPA.


\subsection{Local Surrogate Model and D-Optimal Design}
\label{linear surrogate model and d-optimal design}
After generating a candidate set $\cC$ with size $N=|\cC|$, we aim to identify the best PPA solution.
A naive approach is to evaluate all candidates and pick the best among them.
However, evaluating PPA is expensive.
To reduce the PPA evaluation costs, we learn a local surrogate that linearly maps a candidate's feature vector to its predicted PPA, reducing the number of PPA evaluations. 
Concretely, given a feature map $\phi:\cC\to \RR^d$, we model the PPA composite cost $\SPPA$ of each candidate $c \in\cC$ as  
\begin{align*}
    S_{\mathrm{PPA}}(c)=\phi(c)^\top\theta_\star +\varepsilon(c),
\end{align*}
where $\theta_\star\in \RR^d$ is an underlying model parameter and $\varepsilon:\cC\to \RR$ is the misspecification error.
Although $\SPPA(\cdot)$ is generally nonlinear, our setting is effectively tabular over the finite candidate set induced by our search, hence the name ``local surrogate''.
Because of this, we build a linear model with $N$ dimensions, assigning a dedicated feature to every single candidate. 
This would allow the model to act as a perfect look-up table, memorizing the exact PPA score for each design and resulting in zero misspecification error ($\varepsilon=0$). 
In practice, we trade off feature dimension $d$ and misspecification error $\varepsilon$: smaller $d$ generally reduces the number of expensive PPA evaluations, but may increase $\varepsilon$.
To be more specific, we adopt a set of manually designed features (e.g., normalized cut size, normalized area imbalance; see Section~\ref{sec:feature} for details) with $d\ll N$.

Following ~\cite{Kiefer1960TheEO}, we leverage a result in optimal design theory that shows one is able to fit a ``good enough'' surrogate with a small proportion of the candidate set. 
The intuition is that the amount of independent information available for estimating a surrogate is governed primarily by the feature dimension $d$, rather than the size of candidate set $N$:
In $\RR^d$, any set of linearly independent vectors has a cardinality of at most $d$.
Consequently, large candidate sets residing in $\RR^d$ often contain substantial redundancy when $N\gg d$. 

For example, suppose the candidate set contains $100$ candidates where $\{\phi(c_{i})\}_{i=1}^{99}$ are contained in a small ball $\mathcal B_{\rho}(c_1):=\{b\in \RR^d:\|b-\phi(c_1)\|\le \rho\}$ centered at $\phi(c_1)$ where $\rho>0$ is the radius. The last candidate, $\phi(c_{100})$, is far from this ball: $\min_{x\in \mathcal B_\rho(c_1)}\|x-\phi(c_{100})\|\gg \rho$. 
Because the first 99 candidates are close in the feature space compared to their distance to $\phi(c_{100})$, naturally one would like to construct the coreset $\cK$ to be $\{\phi(c_1),\phi(c_{100})\}$.
We then perform a weighted least squares (WLS) on $\{(\phi(c_1),\SPPA(c_1)),(\phi(c_{100}),\SPPA(c_{100}))\}$  with more weights assigned to $\phi(c_1)$.
We use WLS rather than ordinary least squares because $\phi(c_1)$ and $\phi(c_{100})$ represent varying proportions of the underlying data distribution; weighing $c_1$ more heavily ensures the surrogate model accurately captures the dense region.

Consequently, in order to reduce expensive PPA evaluations, we select a small but informative coreset $\cK:=\{c_j\}_{j=1}^{n}\subseteq \cC$
using D-optimal design \cite{Kiefer1960TheEO}.
The D-optimal design is an optimization problem where the output is a non-negative weight function
\begin{align*}
    \omega:\cC\to [0,1] \qquad \text{such that} \qquad\sum_{i=1}^N \omega (c_i)=1.
\end{align*}
We construct the coreset $\cK$ by keeping the candidates whose corresponding weights exceed a threshold $\mathcal T\ge 0$:
\begin{align*}
    \cK(\mathcal T):=\{c_i: \omega(c_i)> \mathcal T\}\,,
\end{align*}
and then perform weighted least squares to fit a local surrogate that maps the candidate features $\phi(c)$ to the PPA composite cost $\SPPA(c)$:
\begin{align*}
    \hat\theta_{\cK(\cT)}:=\arg\min_{\theta\in \RR^d}\sum_{c\in \mathcal K(\mathcal T)} \omega(c)\cdot \bigg(\phi(c)^\top\theta - \SPPA(c)\bigg)^2.
\end{align*}

The solution to D-optimal design $\omega$ ensures that the resulting WLS solution $\hat\theta_{\cK(\cT)}$ enjoys the following property \cite{Kiefer1960TheEO}:
\begin{align*}
    \max_{c\in \cC}|\SPPA(c) - \phi(c)^\top\hat\theta_{\cK(0)}|\le (1+\sqrt{d})\cE,
\end{align*}
where $\cE=\max_{c\in \cC}|\varepsilon(c)|$.

The above result is a worst-case bound, i.e., no matter how one chooses the candidate set, the resulting WLS solution would never have a misspecification error larger than $(1+\sqrt d)\cE$.
Kiefer and Wolfowitz~\cite{Kiefer1960TheEO} showed that there always exists a coreset $\cK_\star$ such that $|\cK_\star|\le \frac{d(d+1)}{2}$, which is the worst case smallest coreset. 
Finding such a coreset requires solving an exact D-optimal design problem, which is NP-hard \cite{welch1982algorithmic}.
Nevertheless, Our experiments show that the cardinality of the selected coreset is still significantly smaller than that of the candidate set. 
Due to numerical issues, it is rarely the case to obtain a weight function $\omega$ where the weights of the ``discarded'' ones are pushed completely to $0$. Hence, we tune the threshold $\cT$ for selecting the coreset (see Section~\ref{sec:d-opt}).

We calculated the PPA of the candidates in the coreset and fit the surrogate, which is then used to score and rank the full candidate set. 
Since small approximation errors can change the ordering among near-optimal solutions, we do not expect the surrogate's top-1 prediction (i.e., the single candidate it estimates to have the best PPA score) to coincide with the true best candidate. 
We therefore evaluate the surrogate's top-$K$ predictions that are not already covered by the coreset as a verification step (with $K=10$ in our experiments; see Section~\ref{sec:linear_surrogate}).
The candidate with the best PPA score is chosen among all evaluated candidates.

\section{Experiments}
\label{experiments}
We evaluate DOPP on all eight 3D-IC designs provided by Open3DBench. 
Following its methodology, we partition only memory blocks and assign all logic cells to the bottom tier. 
We employ the full Open3DBench 3D-IC flow and PPA evaluation toolchain (i.e., we use the same PDK, configurations, and tools), and use the Open3DBench partitioning algorithm as our baseline. A summary of the designs are
listed in Table~\ref{tab:designs}. 
All experiments were conducted on HPC resources. 
Each evaluation ran on Intel Xeon Gold 6448Y CPU nodes, with an allocation of 4 physical CPU cores and 32 GB of memory.

\begin{table}[t]
    \centering
    \caption{Statistics of the eight designs. Freq.\ represents the clock frequency of the design. Macros are large memory/IP blocks, cells are standard logic gates, and nets are the wire connections between them.}
    \label{tab:designs}
    \begin{tabular}{lcccc}
        \toprule
        Designs & \# Macro & Net & Cell & Freq. (MHz) \\
        \midrule
        ariane133      & 132 & 204K  & 168K  & 250.0  \\
        ariane136      & 136 & 209K  & 171K  & 333.3  \\
        black\_parrot  & 24  & 359K  & 307K  & 500.0  \\
        bp\_be         & 10  & 66K   & 51K   & 1250.0 \\
        bp\_fe         & 11  & 43K   & 33K   & 555.6  \\
        bp\_multi      & 26  & 179K  & 152K  & 1250.0 \\
        bp\_quad       & 220 & 1448K & 1135K & 384.6  \\
        swerv\_wrapper & 28  & 117K  & 98K   & 500.0  \\
        \bottomrule
    \end{tabular}
\end{table}

\subsection{Candidate Set Generation}
\label{sec:candidate}
Fig.~\ref{fig:pareto} illustrates why we adopt a grid-based proxy archive. 
In Fig.~\ref{fig:pareto} (left), the standard multi-objective archive keeps only non-dominated solutions, yielding a relatively small and concentrated proxy frontier. 
Because the estimate of PPA from proxy metrics may not be straightforward, this can be overly restrictive: proxy-dominated solutions may still achieve strong PPA, but would be discarded if we retain only the frontier. 
In Fig.~\ref{fig:pareto} (right), our grid-based archive retains a much more diverse set of proxy-competitive candidates by keeping at most one representative per proxy grid cell, including solutions behind the global Pareto frontier. 
Crucially, after evaluating PPA across this archive, we observe that many of the top-ranked PPA candidates actually reside behind the frontier, which proves our assumption that proxy metrics and true PPA are indeed misaligned. 

To populate this archive, we employ a standard simulated annealing (SA) algorithm as our search algorithm to generate candidates.
As the SA naturally explores the design space, the grid-based archive operates alongside it as a selective filter, preserving the diverse and competitive candidates encountered to construct our candidate set $\cC$.
The grid granularity is a tunable hyper-parameter that directly controls the size of the proxy archive: a finer grid (i.e., smaller cell size) leads to more bins per proxy objective, thereby preserving more candidates.
In Table~\ref{tab:candidate_size}, we vary the grid granularity to generate candidate sets of different sizes. 
Larger candidate sets can increase the chance of containing stronger PPA solutions, but with the same feature construction, they also make a surrogate model more challenging to find the best one, as discussed in Section~\ref{sec:linear_surrogate} (see table~\ref{tab:candidate_size}).

\subsection{Feature Construction}
\label{sec:feature}
For each candidate $c$, we construct a feature vector $\phi(c)\in\RR^d$ for a surrogate model.
The features combine proxy-based partition statistics and hierarchy-aware cohesion signals.
Note that the total feature dimension $d$ varies across different chip designs (as shown in Table~\ref{tab:linear}).
While the number of proxy-based features is fixed across all designs (seven features, $F_1$ to $F_7$), the number of hierarchy cohesion features depends on the number of logical hierarchy clusters present in a specific design. 
In all cases, $d \ll N$.

\textbf{Proxy-based features.} 
Let $cut\_size$ denote the number of inter-tier nets (cut nets), $num\_nets$ the total number of nets, $A_1, A_2$ the memory block areas on the two tiers, and $M_1, M_2$ the numbers of memory blocks on the two tiers.
We use:
\begin{align*}
    F_1=\frac{cut\_size}{num\_nets}, F_2=\frac{|A_1-A_2|}{A_1+A_2}, F_3=\frac{|M_1-M_2|}{M_1+M_2}.
\end{align*}
We further summarize cut-net connectivity by statistics of cut degree:
\begin{align*}
    F_4-F_7 = \{min, max, mean, std\} \text{ of cut-net degree.}
\end{align*}

\textbf{Hierarchy cohesion features.} 
We add hierarchy-aware features to capture structural fragmentation across tiers.
For each logical hierarchy cluster $h$ defined in a netlist, let $S_h$ be the number of logic cells, and let $M_{h,1}$ and $M_{h,2}$ denote the numbers of memory blocks assigned to the lower and upper tiers, respectively, with $M_h=M_{h,1}+M_{h,2}$.
Since logic cells are fixed on the bottom tier, we define the lower-tier cohesion score
\begin{align*}
    coh(h)=\frac{S_hM_{h,1}+\binom{M_{h,1}}{2}}{S_hM_h+\binom{M_h}{2}}.
\end{align*}
Higher values indicate stronger cohesion (less cross-tier fragmentation), while lower values indicate stronger segmentation.

\subsection{Local Surrogate}
\label{sec:linear_surrogate}
Table~\ref{tab:linear} summarizes the performance of the local surrogate using our manually constructed feature map $\phi$. 
Since our goal is to identify the best PPA solution (rather than predict exact PPA values), we report \textbf{Pred@10} which true top-10 PPA candidates are included in the surrogate’s top-10 predictions. 
Exact top-1 prediction is a stringent requirement for an approximate local surrogate. 
This supports our strategy of additionally evaluating the top-$K$ surrogate-predicted candidates.
On these eight designs, the surrogate’s top-1 prediction matches the true top-1 only for \texttt{ariane136} and \texttt{bp\_multi}.
For most designs, the true top-1 is still included in the surrogate’s top-10 predictions, helping to justify our selection of $K=10$.

\begin{table}[h]
    \centering
    \caption{Local surrogate quality on candidate sets. $|\cC|$ is the number of candidates and $d$ is the number of features. For \texttt{swerv\_wrapper}, none of the true top-10 candidates appears in the surrogate top-10; the best of any of the top 10 surrogate-predicted candidates is only the 13th best according to the true PPA.}
    \begin{tabular}{lccc}
         \toprule
         \textbf{Designs} & \textbf{$|\cC|$} & \textbf{$d$} & \textbf{{Pred@10}}\\
         \midrule
         \texttt{ariane133}& 258 & 77  & 1,5,10 \\
        
         \texttt{ariane136}& 152 & 56  & 1,2,3,4,6,8\\
         
         \texttt{black\_parrot}& 248 & 25 & 1 \\
         
         \texttt{bp\_be}& 81 & 15  & 2,3,6,9\\
         
         \texttt{bp\_fe}& 64 & 15 & 1,3,4,6,9\\
         
         \texttt{bp\_multi}& 256 & 25 & 1,8,10 \\
         
         \texttt{bp\_quad}& 492 & 173 & 10 \\
         
         \texttt{swerv\_wrapper}& 263 & 20 & /\\
         \bottomrule
    \end{tabular}
    \label{tab:linear}
\end{table}

Table~\ref{tab:candidate_size} studies how candidate sets' size $|\cC|$ affects surrogate ranking when using the same feature map $\phi$. 
As $|\cC|$ increases, the recovered true ranks in the surrogate top-10 degrade substantially. 
This indicates that, under our current feature map $\phi$, the induced linear model becomes harder to fit accurately over a larger and typically more diverse candidate set, leading to larger ranking errors.
This performance drop indicates that our fixed feature map $\phi$ lacks the expressivity needed to capture the increased diversity of a larger candidate set. Consequently, the linear model underfits, leading to larger ranking errors. 
This trend reflects a limitation of the chosen feature map rather than an inherent limitation of local surrogates: with a more expressive feature map, a linear model could remain accurate even on larger candidate sets.
However, the right-most column of Table~\ref{tab:candidate_size} shows that enlarging $|\cC|$ improves the best PPA solution available for selection.
Together, these results highlight a practical trade-off: larger candidate sets increase the chance of containing better PPA solutions, but they also place higher demands on feature design to maintain reliable surrogate ranking.
\begin{table}[t]
    \centering
    \setlength{\tabcolsep}{4.6pt}
    \caption{Local surrogate quality on different candidate sets size. The last column reports the PPA metrics for the candidate that achieved the best (lowest) PPA composite cost: congestion ($\%$), routed wirelength (m), WNS (ns), TNS (ns), and power (W).}
    \begin{tabular}{lcccc}
         \toprule
         \textbf{Designs} & $|\cC|$ & $d$ & \textbf{\shortstack{Pred@10}} &
         \textbf{\shortstack{Best found PPA in $\cC$}} \\
         \midrule
         \texttt{bp\_multi} & 101 & 25& 1,2,3,4,6,9,10 & {\scriptsize\shortstack{0.1151, 3.43, -6.80, -8061.23, 1.025}}\\
         \texttt{bp\_multi} & 256 & 25& 1,8,10 & {\scriptsize\shortstack{0.1142, 3.40, -6.86, -7672.30, 1.024}}\\
         \texttt{bp\_multi} & 916 & 25& 4 & {\scriptsize\shortstack{0.1117, 3.35, -6.64, -8039.76, 1.021}}\\
         \bottomrule
    \end{tabular}
    \label{tab:candidate_size}
\end{table}

\subsection{D-optimal design}
\label{sec:d-opt}

\begin{table*}[!t]
\centering
\setlength{\tabcolsep}{4.2pt}
\renewcommand{\arraystretch}{1.05}
\caption{Recovery of true top-ranked candidates under different D-optimal design thresholds \(\cT\). For each design, we report the coreset size \(|\cK|\), \textbf{Pred@10} (true top-10 ranks contained in the surrogate top-10 predictions), and \textbf{Eval@10} (true top-10 ranks contained in the evaluated set \(\cK\cup\{\)top-10 surrogate predictions\(\}\)). For \texttt{bp\_quad}, all candidates satisfy \(\omega(c_i) < 8\times10^{-3}\), so no data is reported for \(\cT=10^{-2}\) and \(\cT=8\times10^{-3}\).} 
\begin{tabular}{l c || c c c || c c c || c c c || c c c}
\toprule
\multirow{2}{*}{\textbf{Design}} &
\multirow{2}{*}{$|\cC|$} &
\multicolumn{3}{c}{$\cT=1\times10^{-2}$} &
\multicolumn{3}{c}{$\cT=8\times10^{-3}$} &
\multicolumn{3}{c}{$\cT=5\times10^{-3}$} &
\multicolumn{3}{c}{$\cT=1\times10^{-3}$} \\
\cmidrule(lr){3-5}\cmidrule(lr){6-8}\cmidrule(lr){9-11}\cmidrule(lr){12-14}
& &
$|\cK|$ & \textbf{Pred@10} & \textbf{Eval@10} &
$|\cK|$ & \textbf{Pred@10} & \textbf{Eval@10} &
$|\cK|$ & \textbf{Pred@10} & \textbf{Eval@10} &
$|\cK|$ & \textbf{Pred@10} & \textbf{Eval@10} \\
\midrule
\texttt{ariane133} & 258 & 20 & 1,5 & 1,5 & 45 & / & 1-3,5,10 & 96 & 4 & 1-5,10 & 157 & 1,5,2,10 & 1-6,8-10 \\
\texttt{ariane136} & 152 & 50 & 1 & 1-8 & 63 & 1-4,6 & 1-8 & 92 & 1-4 & 1-8 & 114 & 1-4,6,8 & 1-10 \\
\texttt{black\_parrot}  & 248 & 44 & 10 & 3,8,10 & 57 & 10 & 3,8,10 & 77 & 5,10 & 3,5-7,8,10 & 114 & 1,9,10 & 1,3-6,9-10 \\
\texttt{bp\_be}  & 81  & 32 & 2,3,6,8,10 & 2-4,6,8,10 & 35 & 3,6,10 & 3,4,6,8,10 & 35 & 3,6,10 & 3,4,6,8,10 & 37 & 3,6,10 & 3,4,6,8,10 \\
\texttt{bp\_fe}  & 64  & 36 & 2-6,9 & 1-6,9 & 36 & 2-6,9 & 1-6,9 & 37 & 2-6,9 & 1-6,9 & 39 & 2-6,9 & 1-6,9 \\
\texttt{bp\_multi} & 256 & 44 & 10 & 3,8,10 & 56 & 10 & 3,8,10 & 74 & 5,10 & 3,5-6,8,10 & 102 & 5,10 & 3,5-6,8,10 \\
\texttt{bp\_quad} & 492 & / & / & / & / & / & / & 25 & / & 2 & 348 & 2 & 2-9 \\
\texttt{swerv\_wrapper} & 263 & 44 & / & 6,10 & 51 & / & 6,10 & 60 & / & 6,10 & 73 & / & 6,10 \\
\bottomrule
\end{tabular}
\label{tab:dopt_recovery}
\end{table*}

Table~\ref{tab:dopt_recovery} studies the effect of the D-optimal design threshold 
$\cT$ on coreset size $|\cK|$ and recovery quality. 
\textbf{Pred@10} measures the ranking quality of the surrogate fitted from the D-optimal coreset.
By contrast, \textbf{Eval@10} measures the true top-10 ranks covered by all evaluated candidates, i.e., the union of the coreset and the additionally evaluated top-10 surrogate predictions.
Since the ultimate goal of the framework is to find the best-PPA candidate, \textbf{Eval@10} is the true measure of our framework's success.
We report \textbf{Pred@10} only as a supplementary metric to help understand the surrogate model's underlying performance.

A smaller threshold $\cT$ generally leads to a larger coreset (i.e., the coreset size $|\cK|$ increases).
This is expected, since lowering $\cT$ relaxes the selection criterion and allows more candidates to satisfy the threshold and enter the coreset.

For \textbf{Eval@10}, lower thresholds lead to stronger coverage of the true top-10 set. 
This trend is intuitive because a larger coreset directly evaluates more candidates and therefore has a higher chance of containing strong PPA solutions, and additionally provides more labeled data for fitting the surrogate, which can improve the quality of the additionally evaluated top-10 predictions.
For example, on \texttt{black\_parrot}, the covered true ranks improve from
$3, 8, 10$ at $\cT=1\times10^{-2}$ to almost the full true top-10 at $\cT=1 \times 10^{-3}$.
Similar behavior can also be observed on \texttt{ariane133}, \texttt{ariane136}, \texttt{bp\_multi} and \texttt{bp\_quad}.

For \textbf{Pred@10}, while a lower threshold expands the coreset and provides more training data, its impact on \textbf{Pred@10} is inconsistent.
For example, on \texttt{bp\_be}, the surrogate achieves its strongest prediction recovery at $\cT=1\times10^{-2}$ and become slightly worse as the coreset grows.
A related phenomenon appears on \texttt{bp\_quad}: although fitting on all candidates only recovers rank 10 (see Table~\ref{tab:linear}), the surrogate trained on the D-optimal coreset at $\cT=1\times10^{-3}$ recovers the true rank-2 candidate.
Overall, these cases suggest that \textbf{Pred@10} depends not only on the number of labeled candidates, but also on the feature representation and the specific subset selected for training. Therefore we conclude that enlarging the coreset does not guarantee improved prediction recovery.

We draw two conclusions from the results presented in Table~\ref{tab:dopt_recovery}. 
First, although \textbf{Pred@10} does not admit a monotonic improvement with larger coresets, and in general has no guarantee of perfectly recovering the true top-ranked candidates, the empirical results show that DOPP is still able to select relatively strong PPA solutions from the generated candidate set across all eight designs. 
Second, the quality of the final solution selected by DOPP is positively related to the evaluation budget: as more candidates are evaluated through a larger coreset, \textbf{Eval@10} becomes stronger. This indicates that DOPP can effectively convert additional evaluation compute into improved final PPA performance, which is a major strength of our framework.

\begin{figure}[h]
    \centering
    \includegraphics[width=1.0\linewidth]{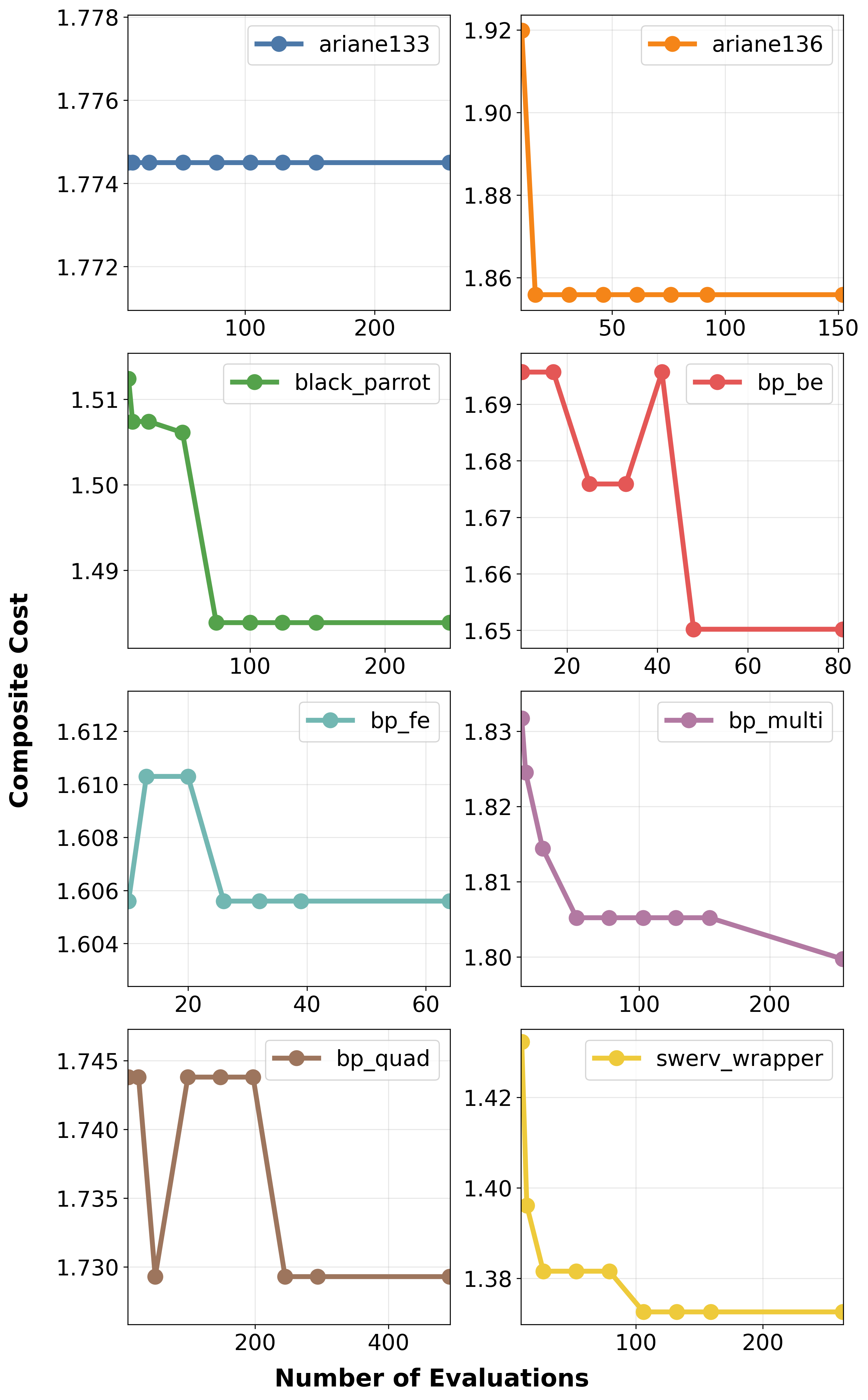}
    \caption{Best PPA found by DOPP versus evaluation budget across eight designs. Each subplot shows the PPA composite cost of the best solution found by DOPP (i.e., the best among all evaluated candidates) as a function of the number of PPA evaluations (x-axis). Lower composite cost indicates better PPA. Each point corresponds to evaluating the top $1\%, 5\%, 10\%, 20\%, 30\%, 40\%, 50\%, 60\%, 100\%$ of candidates, selected by sorting candidates in descending order of their $\omega(c_i)$. We enforce a minimum coreset size of 10, so for smaller designs like \texttt{bp\_fe}, some percentages map to the same evaluation count and fewer distinct points are shown.} 
    \label{fig:budgets}
\end{figure}
Figure~\ref{fig:budgets} further substantiates the second finding above by directly visualizing the relationship between evaluation budget and the best solution obtained by DOPP.
For the majority of the designs, increasing the number of evaluated candidates generally reduces the final PPA composite cost, showing that allocating more evaluations consistently improves the best PPA solution selected from the evaluated set.
A notable exception is \texttt{ariane133}, where the cost curve remains flat; this occurs because our framework successfully identifies the best-performing candidate even within the smallest initial evaluation budget. 

\begin{table*}[!t]
\centering
\renewcommand{\arraystretch}{1.3}
\caption{PPA comparison across eight 3D-IC designs between Open3DBench, an exhaustive oracle that evaluates all candidates (i.e., without D-optimal coreset selection), and DOPP (our framework).
We report congestion (Cong), routed wirelength (rWL), timing (WNS/TNS), and power, together with total CPU time and wall-clock runtime in hours.
``Eval Times'' logs the number of PPA evaluations required.
``Avg Imprv'' summarizes the average relative improvement of DOPP over Open3DBench on the reported PPA metrics.}
\begin{tabular}{|c|c|c|c|c|c|c|c|c|c|c|}
\hline
\textbf{Designs} & \textbf{Methods} & \textbf{\shortstack{Cong\\(\%)}} & \textbf{\shortstack{rWL\\(m)}} & \textbf{\shortstack{WNS\\(ns)}} & \textbf{\shortstack{TNS\\(ns)}} & \textbf{\shortstack{Power\\(W)}} & \textbf{\shortstack{Eval\\Times}} & \textbf{\shortstack{CPU\\time(h)}} & \textbf{\shortstack{Wall\\time(h)}} & \textbf{\shortstack{Avg\\Imprv}}\\
\hline
\multirow{3}{*}{\texttt{ariane133}}
& \textit{Open3DBench} & 0.120 & 5.62 & -1.253 & -2630.37 & 0.358 & 1   & 0.41   & 0.41 & \multirow{3}{*}{8.17\%} \\
& \textit{Exhaustive}  & 0.115 & 5.44 & -1.127 & -2035.89 & 0.355 & 258 & 105.60 & 0.41 & \\
& \textit{DOPP}        & 0.115 & 5.44 & -1.127 & -2035.89 & 0.355 & 20  & 7.98   & 0.41 & \\
\hline
\multirow{3}{*}{\texttt{ariane136}}
& \textit{Open3DBench} & 0.127 & 5.90 & -2.269 & -6122.30 & 0.469 & 1   & 0.47  & 0.47 & \multirow{3}{*}{13.05\%} \\
& \textit{Exhaustive}  & 0.111 & 5.33 & -1.849 & -4793.43 & 0.456 & 152 & 72.23 & 0.49 & \\
& \textit{DOPP}        & 0.111 & 5.33 & -1.849 & -4793.43 & 0.456 & 50  & 23.33 & 0.49 & \\
\hline
\multirow{3}{*}{\texttt{black\_parrot}}
& \textit{Open3DBench} & 0.201 & 7.78 & -5.840 & -3009.20 & 0.374 & 1   & 0.75   & 0.75 & \multirow{3}{*}{12.65\%} \\
& \textit{Exhaustive}  & 0.186 & 7.37 & -5.263 & -1446.89 & 0.371 & 248 & 184.98 & 0.75 & \\
& \textit{DOPP}        & 0.193 & 7.79 & -5.290 & -1522.02 & 0.372 & 44  & 32.86  & 0.75 & \\
\hline
\multirow{3}{*}{\texttt{bp\_be}}
& \textit{Open3DBench} & 0.173 & 2.41 & -0.809 & -93.06   & 0.144 & 1  & 0.15  & 0.15 & \multirow{3}{*}{7.00\%} \\
& \textit{Exhaustive}  & 0.149 & 2.19 & -0.746 & -67.56   & 0.141 & 81 & 12.20 & 0.15 & \\
& \textit{DOPP}        & 0.150 & 2.19 & -0.832 & -80.01   & 0.142 & 32 & 4.82  & 0.15 & \\
\hline
\multirow{3}{*}{\texttt{bp\_fe}}
& \textit{Open3DBench} & 0.127 & 1.30 & -1.314 & -890.39 & 0.283 & 1  & 0.22  & 0.22 & \multirow{3}{*}{7.19\%} \\
& \textit{Exhaustive}  & 0.126 & 1.28 & -1.222 & -659.69 & 0.281 & 64 & 14.22 & 0.23 & \\
& \textit{DOPP}        & 0.126 & 1.28 & -1.222 & -659.69 & 0.281 & 36 & 8.09  & 0.23 & \\
\hline
\multirow{3}{*}{\texttt{bp\_multi}}
& \textit{Open3DBench} & 0.137 & 3.95 & -7.272 & -8975.67 & 1.045 & 1   & 0.24  & 0.24 & \multirow{3}{*}{9.92\%} \\
& \textit{Exhaustive}  & 0.114 & 3.40 & -6.868 & -7672.30 & 1.024 & 256 & 60.23 & 0.24 & \\
& \textit{DOPP}        & 0.113 & 3.39 & -6.547 & -8442.17 & 1.024 & 44  & 10.35 & 0.24 & \\
\hline
\multirow{3}{*}{\texttt{bp\_quad}}
& \textit{Open3DBench} & 0.137 & 40.37 & -1.710 & -26833.40 & 1.837 & 1   & 2.63    & 2.65 & \multirow{3}{*}{10.77\%} \\
& \textit{Exhaustive}  & 0.121 & 36.56 & -1.388 & -20831.30 & 1.818 & 492 & 1294.55 & 2.95 & \\
& \textit{DOPP}        & 0.122 & 36.71 & -1.506 & -21264.40 & 1.816 & 25  & 72.61   & 2.95 & \\
\hline
\multirow{3}{*}{\texttt{swerv\_wrapper}}
& \textit{Open3DBench} & 0.186 & 4.11 & -1.144 & -957.92 & 0.235 & 1   & 0.76   & 0.76 & \multirow{3}{*}{9.25\%} \\
& \textit{Exhaustive}  & 0.155 & 3.43 & -1.138 & -819.66 & 0.232 & 263 & 200.28 & 0.79 & \\
& \textit{DOPP}        & 0.155 & 3.44 & -1.167 & -819.72 & 0.233 & 60  & 33.53  & 0.78 & \\
\hline
\multicolumn{2}{|c|}{\textbf{Avg Imprv}}
& 9.99\% & 7.87\% & 7.75\% & 21.85\% & 1.18\% & \multicolumn{4}{c}{} \\
\cline{1-7}
\end{tabular}
\label{perform_tab}
\end{table*}

\subsection{Overall performance evaluation}

Table~\ref{perform_tab} compares the Open3DBench baseline, and an exhaustive oracle that evaluates all candidates without coreset selection on eight 3D-IC designs, and DOPP.
For all designs except \texttt{bp\_quad}, we set $\mathcal T = 1 \times 10^{-2}$ when constructing the coreset to further reduce the evaluation budget. 
For \texttt{bp\_quad}, we instead use $\mathcal T = 5 \times 10^{-3}$, since all weights are below $1 \times 10^{-2}$. \looseness=-1

Across all designs, DOPP consistently improves PPA over Open3DBench, achieving average gains of $9.99\%$ in Cong, $7.87\%$ in rWL, $7.75\%$ in WNS, $21.83\%$ in TNS, and $1.18\%$ in power. 
Compared with the exhaustive oracle, DOPP achieves near-oracle PPA while requiring substantially fewer PPA evaluations, showing that coreset selection can significantly reduce evaluation cost without compromising solution quality. \looseness=-1

Table~\ref{perform_tab} also highlights a key difference in how PPA evaluations are used.
Prior proxy-driven methods, including Open3DBench, guide the search using proxy objectives rather than PPA, and are not designed to leverage additional PPA evaluations to improve the search or the final selection.
In contrast, DOPP explicitly uses PPA evaluations for training and selection, allocating extra (parallel) PPA evaluations to better identify high-quality solutions.
With sufficient parallel resources, wall-clock runtime remains comparable to the baseline, while total CPU time increases with the number of evaluations.

\begin{table*}[!t]
\centering
\setlength{\tabcolsep}{2.6pt}
\renewcommand{\arraystretch}{1.08}
\caption{Multi-seed comparison (5 seeds) on \texttt{bp\_multi}. Each cell reports the median [min, max] of the final selected solution’s PPA metrics across seeds. For the random baseline, in each seed, we randomly sample and evaluate the exact same number of candidates as DOPP, and then select the candidate with the best overall PPA from that random subset.}
\resizebox{\textwidth}{!}{%
\begin{tabular}{lccccc}
\toprule
\textbf{Method} &
\textbf{Cong (\%)} & \textbf{rWL (m)} & \textbf{WNS (ns)} & \textbf{TNS (ns)} & \textbf{Power (W)} \\
\midrule
\textit{Open3DBench} & 0.134 [0.128, 0.136] & 3.88 [3.76, 3.94] & -7.278 [-8.023, -7.155] & -9104.05 [-10507.5, -8975.67] & 1.040 [1.039, 1.044] \\
\textit{DOPP}   & 0.113 [0.112, 0.113] & 3.35 [3.33, 3.41] & -6.629 [-6.954, -6.546] & -8093.23 [-8156.17, -8069.89] & 1.024 [1.024, 1.026] \\
\textit{Exhaustive} & 0.112 [0.110, 0.113] & 3.31 [3.29, 3.35] & -6.581 [-6.634, -6.482] & -7947.23 [-8039.17, -7850.89] & 1.021 [1.018, 1.021] \\
\textit{Random} & 0.112 [0.110, 0.123] & 3.39 [3.33, 3.48] & -6.851 [-7.054, -6.322] & -8478.23 [-9088.17, -8214.89] & 1.025 [1.019, 1.027] \\
\bottomrule
\end{tabular}%
}

\label{tab:multiseed_doublecol}
\end{table*} 

\begin{figure}[t]
    \centering
    \includegraphics[width=1\linewidth]{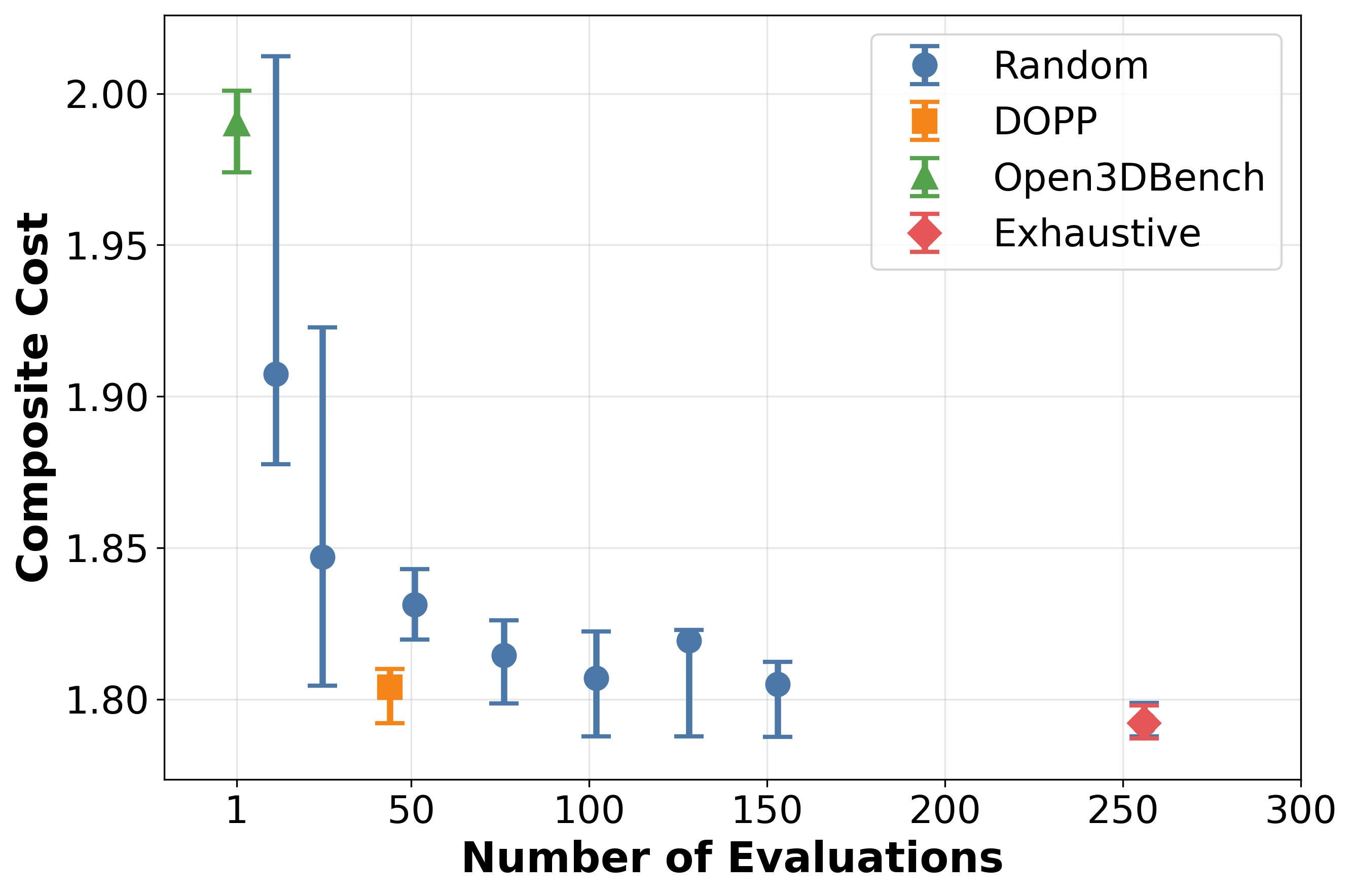}
    \caption{Median composite cost with min-max error bars over 5 seeds on \texttt{multi\_bp}. Each dot shows the median composite cost over 5 random seeds, and each error bar spans the minimum and maximum composite cost across the 5 runs. For the random baseline, the eight points correspond to evaluating randomly sampled subsets containing $5\%, 10\%, 20\%, 30\%, 40\%, 50\%, 60\%, 100\%$ of the full candidate set. The last error bar of the random coincides with the exhaustive. DOPP, Open3DBench, and Exhaustive are shown for comparison.}
    \label{fig:errorbar}
\end{figure}

To further examine robustness under stochasticity, Fig.~\ref{fig:errorbar} and Table~\ref{tab:multiseed_doublecol} report the multiple seeds results on \texttt{bp\_multi}. 
The random baseline exhibits a clear trend: as the number of evaluations increases, its median composite cost gradually improves and its error bars shrink, since evaluating a larger random subset increases the chance of including a high quality candidate and reduces seed to seed variability; at $100\%$ evaluation, it coincides with the exhaustive result by definition. 
However, this also shows that random selection becomes reliable only when the evaluation budget is very large. 
In contrast, DOPP achieves performance close to the exhaustive oracle using a much lower number of evaluations, while maintaining a small performance range across all 5 seeds. 
This indicates that DOPP is robust to the stochasticity introduced by SA in candidate generation, consistently identifying high quality candidates rather than depending on favorable random runs. 
Although the exhaustive oracle also attains strong and stable performance, it requires evaluating the full candidate set, making its computational cost prohibitively expensive. 
Table~\ref{tab:multiseed_doublecol} further shows that this stability on \texttt{bp\_multi} also carries over to the underlying PPA metrics across all 5 seeds, providing additional evidence of DOPP’s robustness.

\begin{table}[h]
\centering
\footnotesize 
\setlength{\tabcolsep}{3pt} 
\renewcommand{\arraystretch}{1.08}
\caption{Comparison under different user-defined final PPA preferences on \texttt{ariane133}. The timing-focused case only considers \textbf{WNS (ns)} and \textbf{TNS (ns)}, the routing-focused case only considers \textbf{Cong (\%)} and \textbf{rWL (m)}, the power-focused case only considers \textbf{power (W)}, and the balanced case considers all PPA metrics jointly. Open3DBench does not support directly adjusting the final PPA preferences to retrieve the corresponding solutions, whereas DOPP can adapt its final selection accordingly.}
\begin{tabular}{l l ccccc}
\toprule
\textbf{Preference} & \textbf{Method} & \textbf{Cong} & \textbf{rWL} & \textbf{WNS} & \textbf{TNS} & \textbf{Power} \\
\midrule
\multirow{2}{*}{Timing}  & Open3DBench & 0.120 & 5.62 & -1.253 & -2630.37 & 0.358 \\
                         & DOPP        & 0.114 & 5.44 & -1.127 & -2035.89 & 0.354 \\
\cmidrule(lr){1-7}
\multirow{2}{*}{Routing} & Open3DBench & 0.120 & 5.62 & -1.253 & -2630.37 & 0.358 \\
                         & DOPP        & 0.110 & 5.27 & -1.231 & -2570.80 & 0.355 \\
\cmidrule(lr){1-7}
\multirow{2}{*}{Power}   & Open3DBench & 0.120 & 5.62 & -1.253 & -2630.37 & 0.358 \\
                         & DOPP        & 0.114 & 5.45 & -1.123 & -2358.42 & 0.349 \\
\cmidrule(lr){1-7}
\multirow{2}{*}{Balanced}& Open3DBench & 0.120 & 5.62 & -1.253 & -2630.37 & 0.358 \\
                         & DOPP        & 0.115 & 5.44 & -1.127 & -2035.89 & 0.355 \\
\bottomrule
\end{tabular}
\label{tab:user_pref}
\end{table}

Table~\ref{tab:user_pref} highlights an important strength of DOPP: it supports final solution selection according to user-preferred PPA metrics. 
Because DOPP is a PPA-driven framework, it naturally enables preference-based retrieval in the final PPA space. 
This is particularly useful in practice when different designers prioritize different objectives, such as timing, routing quality, or power (depending on downstream design needs).

In contrast, proxy-driven methods such as Open3DBench do not directly optimize for user-specified final PPA preferences. 
Although such methods can bias solution generation by changing the weights of their internal proxy objectives, it is difficult to translate a desired final-PPA target into the appropriate proxy weights because the relationship between proxies and final PPA is indirect and often mismatched. 
For example, Open3DBench selects solutions based on fixed weights over proxies such as cut size and area imbalance. 
If a designer instead wants a timing-preferred or power-preferred solution, it is not clear how the weights between cut size and area imbalance should be adjusted to reliably realize that final PPA preference. 
In practice, this typically requires repeated attempts guided by heuristics or human experience.
In contrast, DOPP can directly select the corresponding solution according to the user’s desired PPA trade-off, without relying on proxy-weight tuning.

As shown on \texttt{ariane133} (Table \ref{tab:user_pref}), DOPP successfully retrieves solutions optimized for the user's priority whether that is timing, routing, power, or a balanced improvement across all metrics. 
Crucially, for any given preference, DOPP selects the single best candidate tailored to that goal, rather than outputting a batch of trade-off candidates. 
This confirms that DOPP seamlessly adapts to diverse, user-defined design preferences.

\section{Conclusion}
DOPP is a PPA-driven selection framework for 3D-IC netlist partitioning that closes the gap between proxy-based search and expensive final PPA evaluation. 
Starting from a proxy-diverse candidate set, DOPP uses D-optimal design to select an informative set of candidates to evaluate, fits a local surrogate to rank the remaining candidates, and verifies a small set of top predictions before selecting the final solution.

Across eight Open3DBench designs, DOPP consistently improves final PPA over the Open3DBench baseline while maintaining comparable wall-clock runtime through parallel evaluation. 
Compared with exhaustive evaluation, DOPP achieves comparable solution quality while evaluating only a small fraction of the candidate set, substantially reducing evaluation cost.
Additional experiments further show that increasing the evaluation budget generally improves the final selected solution, demonstrating that DOPP can effectively translate extra expensive evaluations into better PPA outcomes.

Because DOPP operates directly in the true PPA space, it also supports flexible final selection under different designer preferences over congestion, routed wirelength, timing, and power.
More broadly, this work shows that backend PPA evaluation need not remain a passive end-of-flow validation step. 
When used selectively and systematically, it can serve as an effective optimization signal for practical, budget-aware 3D-IC design.

Despite these gains, DOPP remains dependent on the quality of the generated candidate set and the expressiveness of the manually designed features used by the local surrogate; performance drops as the candidate set grows in size and diversity. 
Future work will investigate richer surrogate representations, more adaptive surrogate modeling strategies, and extensions to broader 3D-IC partitioning scenarios.

\section*{Acknowledgment}
Part of this work has taken place in the Intelligent Robot Learning (IRL) Lab at the University of Alberta, which is supported in part by research grants from Alberta Innovates; Alberta Machine Intelligence Institute (Amii); a Canada CIFAR AI Chair, Amii; Digital Research Alliance of Canada; Mitacs; and the National Science and Engineering Research Council (NSERC).

\bibliographystyle{unsrt} 
\bibliography{references}

\end{document}